\documentclass[10pt,twocolumn,letterpaper]{article}
\usepackage{cvpr}
\usepackage{times}
\usepackage{epsfig}
\usepackage{graphicx}
\usepackage{amsmath}
\usepackage{amssymb}
\usepackage{pbox}
\usepackage{epstopdf}
\usepackage{subfigure}
\usepackage{xspace}
\usepackage{comment}

\newcommand{\cmt}[2]{[#1: #2]}

\newcommand{\MATTHIAS}[1]{\cmt{{\bf Matthias}}{{\bf \color{red} #1}}}

\newcommand{\DCNN}{volumetric CNN\xspace}
\newcommand{\DCNNs}{volumetric CNNs\xspace}

\newcommand{\MVCNN}{multi-view CNN\xspace}
\newcommand{\MVCNNs}{multi-view CNNs\xspace}

\newcommand{\AbbrMVCNNs}{MVCNNs\xspace}
\newcommand{\MOVCNN}{MO-VCNN\xspace}

\newcommand{\SuMVCNN}{Su-MVCNN\xspace}

\newcommand\blfootnote[1]{%
	\begingroup
	\renewcommand\thefootnote{}\footnote{#1}%
	\addtocounter{footnote}{-1}%
	\endgroup
}

\newcommand{\mypara}{\vspace*{-10pt}\paragraph}
\usepackage[pagebackref=true,breaklinks=true,letterpaper=true,colorlinks,bookmarks=false]{hyperref}

\cvprfinalcopy 
\def\cvprPaperID{1294} 

\ifcvprfinal\fi
\begin{document}

\title{Volumetric and Multi-View CNNs for Object Classification on 3D Data}

\author{First Author\\
Institution1\\
Institution1 address\\
{\tt\small firstauthor@i1.org}
\and
Second Author\\
Institution2\\
First line of institution2 address\\
{\tt\small secondauthor@i2.org}
}

\author{Charles R. Qi$^*$~~~~~Hao Su$^*$~~~~~Matthias Nie{\ss}ner~~~~~Angela Dai~~~~~Mengyuan Yan~~~~~Leonidas J. Guibas \vspace{0.1cm} \\ 
Stanford University
}
	
\maketitle

\begin{abstract}


3D shape models are becoming widely available and easier to capture, making available 3D information crucial for progress in object classification.
Current state-of-the-art methods rely on CNNs to address this problem. 
Recently, we witness two types of CNNs being developed: CNNs based upon volumetric representations versus CNNs based upon multi-view representations. 
Empirical results from these two types of CNNs exhibit a large gap, indicating that existing \DCNN architectures and approaches are unable to fully exploit the power of 3D representations. 
In this paper, we aim to improve both \DCNNs and \MVCNNs according to extensive analysis of existing approaches. 
To this end, we introduce two distinct network architectures of \DCNNs.
In addition, we examine \MVCNNs, where we introduce multi-resolution filtering in 3D.
Overall, we are able to outperform current state-of-the-art methods for both \DCNNs and \MVCNNs. 
We provide extensive experiments designed to evaluate underlying design choices, thus providing a better understanding of the space of methods available for object classification on 3D data.

\end{abstract}

\section{Introduction}\label{sec:intro}
\blfootnote{* indicates equal contributions.}


Understanding 3D environments is a vital element of modern computer vision research due to paramount relevance in many vision systems, spanning a wide field of application scenarios from self-driving cars to autonomous robots.
Recent advancements in real-time SLAM techniques and crowd-sourcing of virtual 3D models have additionally facilitated the availability of 3D data. \cite{silberman2012indoor,xiao2013sun3d,song2015sun,wu20153d,chang2015shapenet}.
This development has encouraged the lifting of 2D to 3D for deep learning, opening up new opportunities with the additional information of 3D data; e.g., aligning models is easier in 3D Euclidean space. 
In this paper, we specifically focus on the object classification task on 3D data obtained from both CAD models and commodity RGB-D sensors.
In addition, we demonstrate retrieval results in the supplemental material.


While the extension of 2D convolutional neural networks to 3D seems natural, the additional
computational complexity (volumetric domain) and data sparsity introduces significant challenges; for instance, in an image, every pixel contains observed information, whereas in 3D, a shape is only defined on its surface.
Seminal work by Wu et al. \cite{wu20153d} propose \DCNN architectures on volumetric grids for object classification and retrieval.
While these approaches achieve good results, it turns out that training a CNN on multiple 2D views achieves a significantly higher performance, as shown by Su et al. \cite{su15mvcnn}, who augment their 2D CNN with pre-training from ImageNet RGB data \cite{deng2009imagenet}.
These results indicate that existing 3D CNN architectures and approaches are unable to fully exploit the power of 3D representations.
In this work, we analyze these observations and evaluate the design choices.
Moreover, we show how to reduce the gap between \DCNNs and \MVCNNs by efficiently augmenting training data, introducing new CNN architectures in 3D.
Finally, we examine \MVCNNs; our experiments show that we are able to improve upon state of the art with improved training data augmentation and a new multi-resolution component.

\paragraph{Problem Statement}
We consider volumetric representations of 3D point clouds or meshes as input to the 3D object classification problem. 
This is primarily inspired by recent advances in real-time scanning technology, which use volumetric data representations.
We further assume that the input data is already pre-segmented by 3D bounding boxes. 
In practice, these bounding boxes can be extracted using the sliding windows, object proposals, or background subtraction. 
The output of the method is the category label of the volumetric data instance.

\paragraph{Approach}
We provide a detailed analysis over factors that influence the performance of \DCNNs, including network architecture and volumn resolution. Based upon our analysis, we strive to improve the performance of \DCNNs. 
We propose two \DCNN network architectures that signficantly improve state-of-the-art of volumetric CNNs on 3D shape classification. \emph{This result has also closed the gap between \DCNNs and \MVCNNs}, when they are provided with 3D input discretized at $30\times 30\times 30$ 3D resolution. 
The first network introduces auxiliary learning tasks by classifying part of an object, which help to scrutize details of 3D objects more deeply.
The second network uses long anisotropic kernels to probe for long-distance interactions. Combining data augmentation with a multi-orientation pooling, we observe significant performance improvement for both networks.
We also conduct extensive experiments to study the influence of volume resolution, which sheds light on future directions of improving \DCNNs. 

Furthermore, we introduce a new multi-resolution component to \MVCNNs, which improves their already compelling performance.

In addition to providing extensive experiments on 3D CAD model datasets, we also introduce a dataset of real-world 3D data, constructed using dense 3D reconstruction taken with \cite{niessner2013real}. Experiments show that our networks can better adapt from synthetic data to this real-world data than previous methods.




\section{Related Work} \label{sec:prev_work}

\paragraph{Shape Descriptors}
A large variety of shape descriptors has been developed in the computer vision and graphics community.
For instance, shapes can be represented as histograms or bag-of-feature models which are constructed from surface normals and curvatures \cite{horn1984extended}.
Alternatives include models based on distances, angles, triangle areas, or tetrahedra volumes \cite{osada2002shape}, local shape diameters measured at densely-sampled surface points \cite{chaudhuri2010data}, Heat kernel signatures \cite{bronstein2011shape,kokkinos2012intrinsic}, or extensions of SIFT and SURF feature descriptors to 3D voxel grids \cite{knopp2010hough}.
The spherical harmonic descriptor (SPH) \cite{kazhdan2003rotation} and the Light Field descriptor (LFD) \cite{chen2003visual} are other popular descriptors.
LFD extracts geometric and Fourier descriptors from object silhouettes rendered from several different viewpoints, and can be directly applied to the shape classification task.
In contrast to recently developed feature learning techniques, these features are hand-crafted and do not generalize well across different domains.

\begin{figure}[t!]
	\centering
	\includegraphics[width=\linewidth]{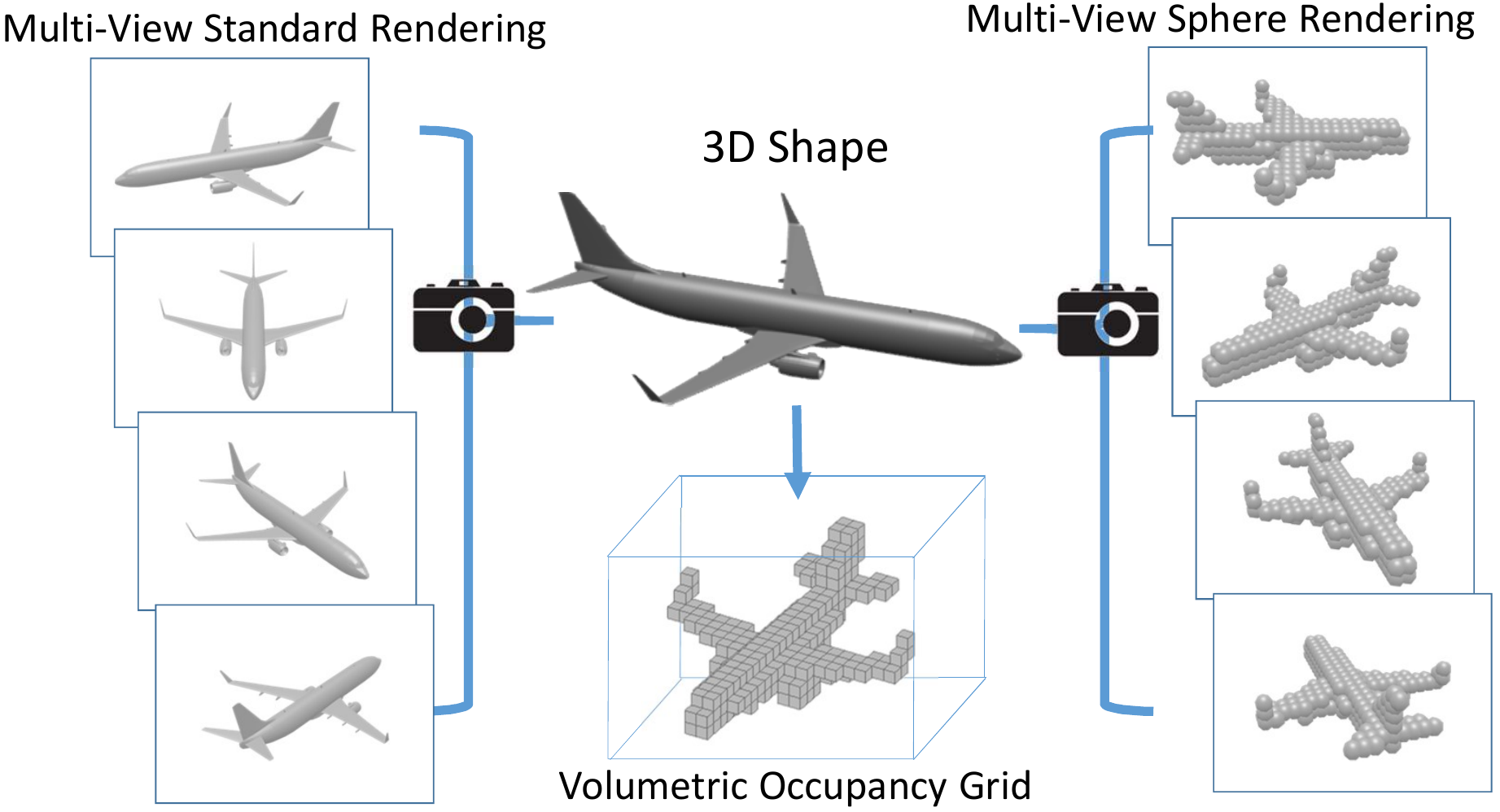}
	\caption{3D shape representations. }
	\label{fig:data_representation}
\end{figure}
\paragraph{Convolutional Neural Networks}

Convolutional Neural Networks (CNNs) \cite{lecun1998gradient} have been successfully used in different areas of computer vision and beyond.
In particular, significant progress has been made in the context of learning features.
It turns out that training from large RGB image datasets  (e.g., ImageNet \cite{deng2009imagenet}) is able to learn general purpose image descriptors that outperform hand-crafted features for a number of vision tasks, including object detection, scene recognition, texture recognition and classification \cite{donahue2013decaf,girshick2014rich,razavian2014cnn,cimpoi2014describing,han2015matchnet}.
This significant improvement in performance on these tasks has decidedly moved the field forward.


\paragraph{CNNs on Depth and 3D Data}

With the introduction of commodity range sensors, the depth channel became available to provide additional information that could be incorporated into common CNN architectures.
A very first approach combines convolutional and recursive neural networks for learning features and classifying RGB-D images \cite{socher2012convolutional}.
Impressive performance for object detection from RGB-D images has been achieved using a geocentric embedding for depth images that encodes height above ground and angle with gravity for each pixel in addition to the horizontal disparity \cite{gupta2014learning}.
Recently, a CNN architecture has been proposed where the RGB and depth data are processed in two separate streams; in the end, the two streams are combined with a late fusion network \cite{eitel15iros}.
All these descriptors operate on single RGB-D images, thus processing 2.5D data.

Wu et al. \cite{wu20153d} lift 2.5D to 3D with their 3DShapeNets approach by  categorizing each voxel as free space, surface or occluded, depending on whether it is in front of, on, or behind the visible surface (i.e., the depth 
value) from the depth map.
The resulting representation is a 3D binary voxel grid, which is the input to a CNN with 3D filter banks.
Their method is particularly relevant in the context of this work, as they are the first to apply CNNs on a 3D representation.
A similar approach is VoxNet \cite{maturana2015voxnet}, which also uses  binary voxel grids and a corresponding 3D CNN architecture.
The advantage of these approaches is that it can process different sources of 3D data, including LiDAR point clouds, RGB-D point clouds, and CAD models; we likewise follow this direction.

An alternative direction is to exploit established 2D CNN architectures; to this end, 2D data is extracted from the 3D representation.
In this context, DeepPano \cite{shi2015deeppano} converts 3D shapes into panoramic views;  i.e., a cylinder projection around its principle axis.
Current state-of-the-art uses multiple rendered views, and trains a CNN that can process all views jointly \cite{su15mvcnn}.
This multi-view CNN (MVCNN) is pre-trained on ImageNet \cite{deng2009imagenet} and uses view-point pooling to combine all streams obtained from each view.
A similar idea on stereo views has been proposed earlier \cite{lecun2004learning}.

\section{Analysis of state-of-the-art 3D Volumetric CNN versus Multi-View CNN} \label{sec:diagnosis}


 \begin{figure}[h]
 	\centering
 	\includegraphics[width=\linewidth]{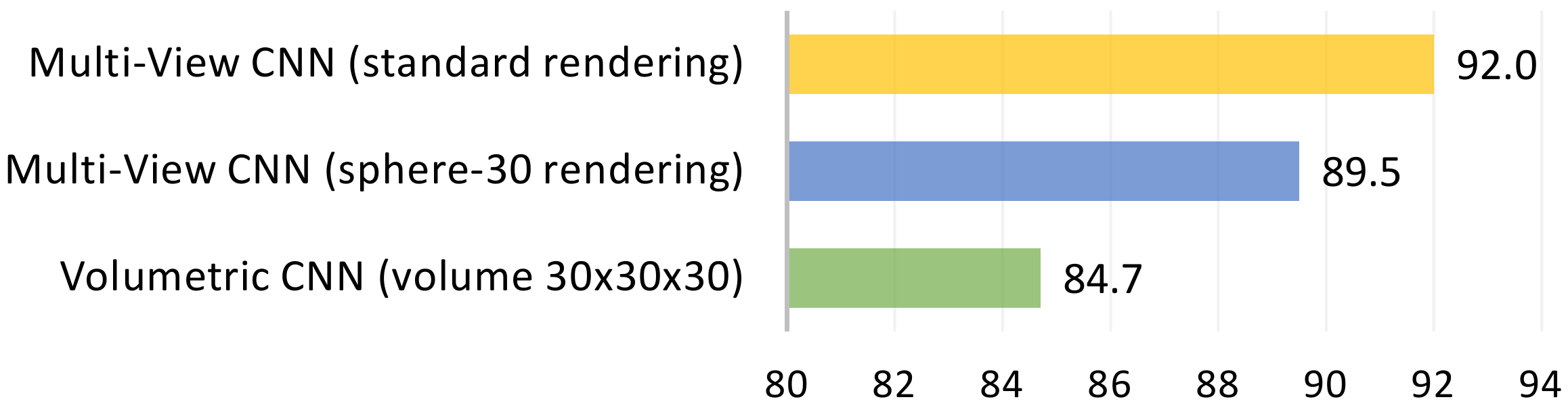}
 	\caption{Classification accuracy.  Yellow and blue bars:  Performance drop of \MVCNN due to discretization of CAD models in rendering. Blue and green bars: Volumetric CNN is significantly worse than \MVCNN, even though their inputs have similar amounts of information. This indicates that the network of the \DCNN is weaker than that of the \MVCNN.}
 	\label{fig:vcnn_vs_mvcnn}
 \end{figure}
 
Two representations of generic 3D shapes are popularly used for object classification, volumetric and multi-view (Fig~\ref{fig:data_representation}). The volumetric representation encodes a 3D shape as a 3D tensor of binary or real values. The multi-view representation encodes a 3D shape as a collection of renderings from multiple viewpoints. Stored as tensors, both representations can easily be used to train convolutional neural networks, i.e., \DCNNs and \MVCNNs. 

Intuitively, a volumetric representation should encode as much information, if not more, than its multi-view counterpart. However, experiments indicate that \MVCNNs produce superior performance in object classification. Fig~\ref{fig:vcnn_vs_mvcnn} reports the classification accuracy on the ModelNet40 dataset by state-of-the-art volumetric/multi-view architectures\footnote{We train models by replicating the architecture of \cite{wu20153d} for \DCNNs and \cite{su15mvcnn} for \MVCNNs. All networks are trained in an end-to-end fashion. All methods are trained/tested on the same split for fair comparison. The reported numbers are average instance accuracy. See Sec~\ref{sec:experiments} for details.}. A \DCNN based on voxel occupancy (green) is $7.3\%$ worse than a \MVCNN (yellow). 

We investigate this performance gap in order to ascertain how to improve \DCNNs. The gap seems to be caused by two factors: input resolution and network architecture differences. The \MVCNN down-samples each rendered view to $227\times 227$ pixels (Multi-view Standard Rendering in Fig~\ref{fig:data_representation}); to maintain a similar computational cost, the \DCNN uses a $30\times 30\times 30$ occupancy grid  (Volumetric Occupancy Grid in Fig~\ref{fig:data_representation})\footnote{Note that $30\times 30 \times 30 \approx 227 \times 227$.}. As shown in Fig~\ref{fig:data_representation}, the input to the \MVCNN captures more detail. 

However, the difference in input resolution is not the primary reason for this performance gap, as evidenced by further experiments.  We compare the two networks by providing them with data containing similar level of detail. To this end, we feed the \MVCNN with renderings of the $30\times 30\times 30$ occupancy grid using \emph{sphere rendering}\footnote{It is computationally prohibitive to match the \DCNN resolution to \MVCNN, which would be $227\times227\times 227$.}, i.e., for each occupied voxel, a ball is placed at its center, with radius equal to the edge length of a voxel (Multi-View Sphere Rendering in Fig~\ref{fig:data_representation}). We train the \MVCNN from scratch using these sphere renderings.
The accuracy of this \MVCNN is reported in blue.
 
\begin{figure*}[t!]
	\centering
	\includegraphics[width=0.8\linewidth]{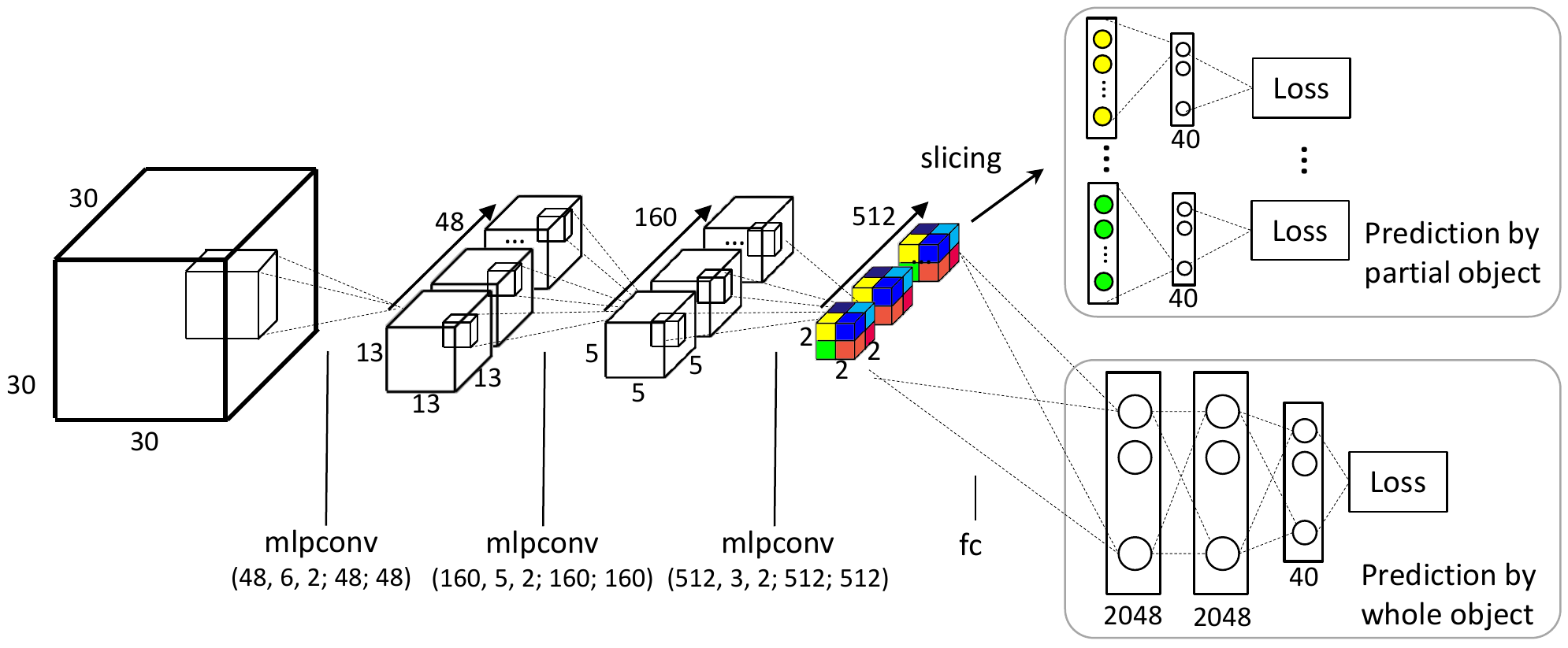}
	\caption{Auxiliary Training by Subvolume Supervision (Sec~\ref{sec:auxiliary_training}). The main innovation is that we add auxiliary tasks to predict class labels that focus on part of an object, intended to drive the CNN to more heavily exploit local discriminative features. An mlpconv layer is a composition of three conv layers interleaved by ReLU layers. The five numbers under mlpconv are the number of channels, kernel size and stride of the first conv layer, and the number of channels of the second and third conv layers, respectively. The kernel size and stride of the second and third conv layers are $1$. For example, $\text{mlpconv}(48,6,2; 48; 48)$ is a composition of $\text{conv}(48, 6, 2)$, ReLU, $\text{conv}(48, 1, 1)$, ReLU, $\text{conv}(48, 1, 1)$ and ReLU layers.  Note that we add dropout layers with rate=$0.5$ after fully connected layers.}
	\label{fig:auxiliary_training}
\end{figure*}

As shown in Fig~\ref{fig:vcnn_vs_mvcnn}, even with similar level of object detail, the \DCNN (green) is $4.8\%$ worse than the \MVCNN (blue). That is, \emph{there is still significant room to improve the architecture of \DCNNs}. This discovery motivates our efforts in Sec~\ref{sec:DCNN} to improve \DCNNs. Additionally, low-frequency information in 3D seems to be quite discriminative for object classification---it is possible to achieve $89.5\%$ accuracy (blue) at a resolution of only $30\times 30\times 30$. This discovery motivates our efforts in Sec~\ref{sec:mvcnn} to improve \MVCNNs with a 3D multi-resolution approach.


\section{Volumetric Convolutional Neural Networks} \label{sec:vcnn}
\label{sec:DCNN}
\subsection{Overview}

We improve \DCNNs through three separate means: 1) introducing new network structures; 2) data augmentation; 3) feature pooling.

\paragraph{Network Architecture} 

We propose two network variations that significantly improve state-of-the-art CNNs on 3D volumetric data. 
The first network is designed to mitigate overfitting by introducing auxiliary training tasks, which are themselves challenging. These auxiliary tasks encourage the network to predict object class labels from partial subvolumes. Therefore, no additional annotation efforts are needed.  
The second network is designed to mimic \MVCNNs, as they are strong in 3D shape classification. Instead of using rendering routines from computer graphics, our network projects a 3D shape to 2D by convolving its 3D volume with an anisotropic probing kernel. This kernel is capable of encoding long-range interactions between points. An image CNN is then appended to classify the 2D projection. Note that the training of the projection module and the image classification module is end-to-end. This emulation of \MVCNNs achieves similar performance to them, using only standard layers in CNN. 

In order to mitigate overfitting from too many parameters, we adopt the mlpconv layer from \cite{lin2013network} as our basic building block in both network variations.


\paragraph{Data Augmentation} Compared with 2D image datasets, currently available 3D shape datasets are limited in scale and variation. To fully exploit the design of our networks, we augment the training data with different azimuth and elevation rotations. 
This allows the first network to cover local regions at different orientations, and the second network to relate distant points at different relative angles.

\paragraph{Multi-Orientation Pooling} Both of our new networks are sensitive to shape orientation, i.e., they capture different information at different orientations. To capture a more holistic sense of a 3D object, we add an orientation pooling stage that aggregates information from different orientations.

\begin{figure*}[t!]
	\centering
	\includegraphics[width=0.8\linewidth]{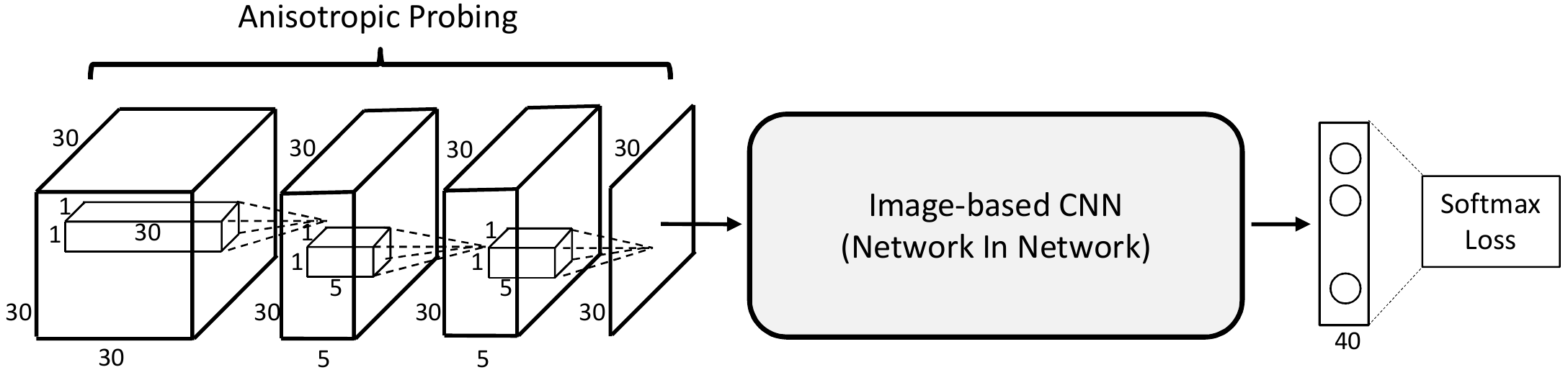}
	\caption{CNN with Anisotropic Probing kernels. We use an elongated kernel to convolve the 3D cube and aggregate information to a 2D plane. Then we use a 2D NIN (NIN-CIFAR10~\cite{lin2013network}) to classify the 2D projection of the original 3D shape. 
		\label{fig:anisotropic_probing}}
\end{figure*}

\subsection{Network 1: Auxiliary Training by Subvolume Supervision}
\label{sec:auxiliary_training}
We observe significant overfitting when we train the \DCNN proposed by \cite{wu20153d} in an end-to-end fashion (see supplementary). When the \DCNN overfits to the training data, it has no incentive to continue learning. We thus introduce auxiliary tasks that are closely correlated with the main task but are difficult to overfit, so that learning continues even if our main task is overfitted.  

These auxiliary training tasks also predict the same object labels, but the predictions are made solely on a local subvolume of the input.
Without complete knowledge of the object, the auxiliary tasks are more challenging, and can thus better exploit the discriminative power of local regions. This design is different from the classic multi-task learning setting of hetergenous auxiliary tasks, which inevitably requires collecting additional annotations (e.g., conducting both object classification and detection \cite{girshick2015fast}). 

We implement this design through an architecture shown in Fig~\ref{fig:auxiliary_training}. 
The first three layers are mlpconv (multilayer perceptron convolution) layers, a 3D extension of the 2D mlpconv proposed by \cite{lin2013network}. 
The input and output of our mlpconv layers are both 4D tensors. 
Compared with the standard combination of linear convolutional layers and max pooling layers, mlpconv has a three-layer structure and is thus a universal function approximator if enough neurons are provided in its intermediate layers. Therefore, mlpconv is a powerful filter for feature extraction of local patches, enhancing approximation of more abstract representations. In addition, mlpconv has been validated to be more discriminative with fewer parameters than ordinary convolution with pooling~\cite{lin2013network}. 

At the fourth layer, the network branches into two. 
The lower branch takes the whole object as input for traditional classification. 
The upper branch is a novel branch for auxiliary tasks. 
It slices the $512\times 2\times 2 \times 2$ 4D tensor ($2$ grids along $x$, $y$, $z$ axes and $512$ channels) into $2\times 2\times 2=8$ vectors of dimension $512$. We set up a classification task for each vector. 
A fully connected layer and a softmax layer are then appended independently to each vector to construct classification losses. Simple calculation shows that the receptive field of each task is $22\times 22\times 22$, covering roughly $2/3$ of the entire volume.


\subsection{Network 2: Anisotropic Probing}
\label{sec:anisotropic_probing}
The success of \MVCNNs is intriguing. \MVCNNs first project 3D objects to 2D and then make use of well-developed 2D image CNNs for classification. Inspired by its success, we design a neural network architecture that is also composed of the two stages. However, while \MVCNNs use external rendering pipelines from computer graphics, we achieve the 3D-to-2D projection using network layers in a manner similar to `X-ray scanning'. 

Key to this network is the use of an elongated anisotropic kernel which helps capture the global structure of the 3D volume.
As illustrated in Fig~\ref{fig:anisotropic_probing}, the neural network has two modules: an anisotropic probing module and a network in network module. 
The anisotropic probing module contains three convolutional layers of elongated kernels, each followed by a nonlinear ReLU layer. 
Note that both the input and output of each layer are 3D tensors. 

In contrast to traditional isotropic kernels, an anisotropic probing module has the advantage of aggregating long-range interactions in the early feature learning stage with fewer parameters. 
As a comparison, with traditional neural networks constructed from isotropic kernels, introducing long-range interactions at an early stage can only be achieved through large kernels, which inevitably introduce many more parameters.
After anisotropic probing, we use an adapted NIN network \cite{lin2013network} to address the classification problem.

Our anistropic probing network is capable of capturing internal structures of objects through its X-ray like projection mechanism. This is an ability not offered by standard rendering. 
Combined with multi-orientation pooling (introduced below), it is possible for this probing mechanism to capture any 3D structure, due to its relationship with the Radon transform. 

In addition, this architecture is scalable to higher resolutions, since all its layers can be viewed as 2D. While 3D convolution involves computation at locations of cubic resolution, we maintain quadratic compute.

 \begin{figure}[b!]
 	\centering
 	\includegraphics[width=0.8\linewidth]{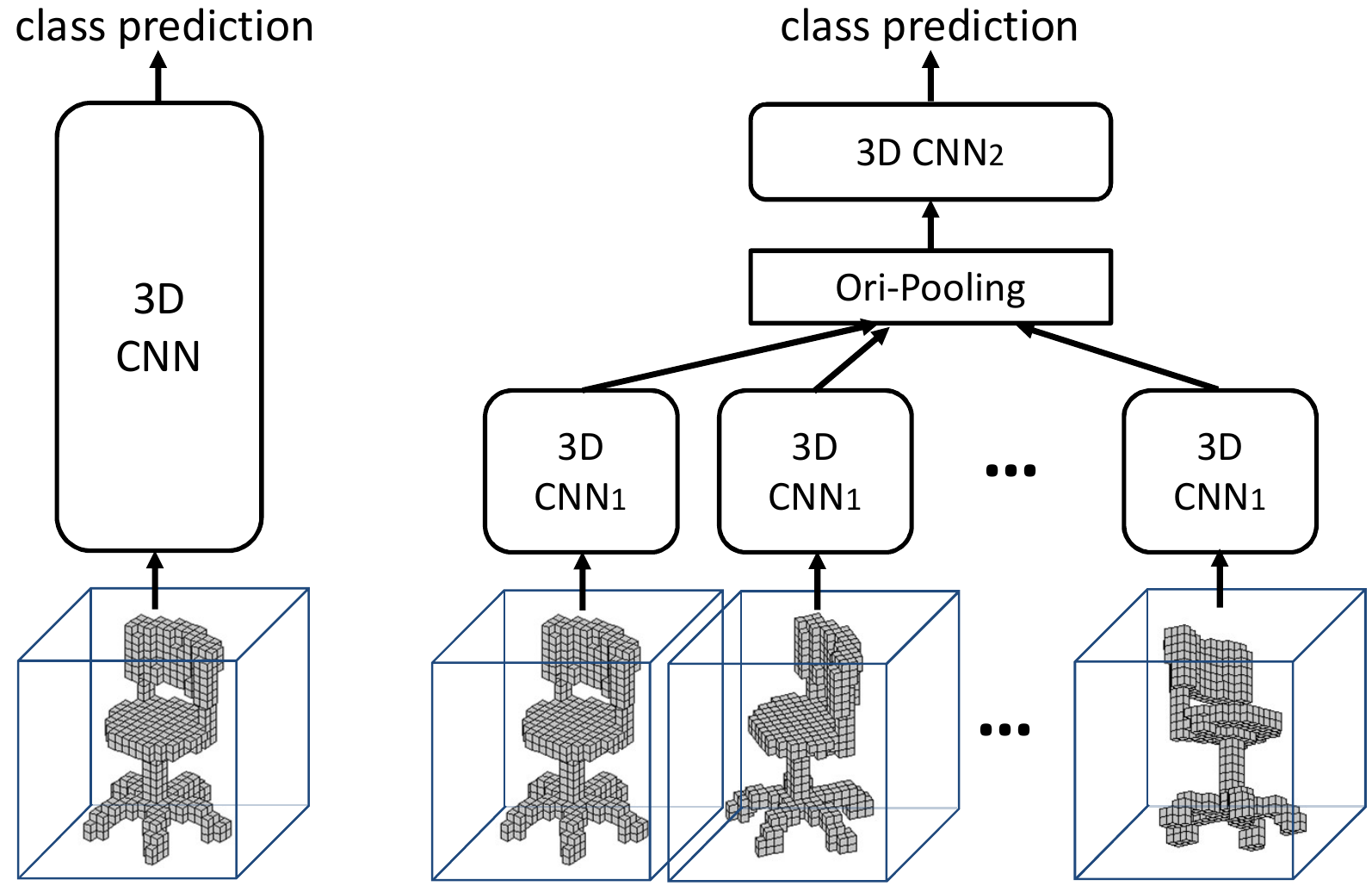}
 	\caption{Left: Volumetric CNN (single orientation input). Right: Multi-orientation volumetric CNN (\MOVCNN), which takes in various orientations of the 3D input, extracts features from shared $\text{CNN}_1$ and then pass pooled feature through another network $\text{CNN}_2$ to make a prediction.}
 	\label{fig:multi_ori}
 \end{figure}
\subsection{Data Augmentation and Multi-Orientation Pooling} \label{sec:arch:aug}
The two networks proposed above are both sensitive to model orientation. In the subvolume supervision method, different model orientations define different local subvolumes; in the anisotropic probing method, only voxels of the same height and along the probing direction can have interaction in the early feature extraction stage. 
Thus it is helpful to augment the training data by varying object orientation and combining predictions through orientation pooling. 

Similar to \SuMVCNN~\cite{su15mvcnn} which aggregates information from multiple view inputs through a view-pooling layer and follow-on fully connected layers, we sample 3D input from different orientations and aggregate them in a multi-orientation volumetric CNN (\MOVCNN) as shown in Fig~\ref{fig:multi_ori}. At training time, we generate different rotations of the 3D model by changing both azimuth and elevation angles, sampled randomly. A volumetric CNN is firstly trained on single rotations. Then we decompose the network to $\text{CNN}_1$ (lower layers) and $\text{CNN}_2$ (higher layers) to construct a multi-orientation version. The \MOVCNN's weights are initialized by a previously trained volumetric CNN with $\text{CNN}_1$'s weights fixed during fine-tuning. While a common practice is to extract the highest level features (features before the last classification linear layer) of multiple orientations, average/max/concatenate them, and train a linear SVM on the combined feature, this is just a special case of the \MOVCNN.

Compared to~3DShapeNets~\cite{wu20153d} which only augments data by rotating around vertical axis, our experiment shows that orientation pooling combined with elevation rotation can greatly increase performance.



\section{Multi-View Convolutional Neural Networks} \label{sec:mvcnn}
The \MVCNN proposed by \cite{su15mvcnn} is a strong alternative to volumetric representations. 
This multi-view representation is constructed in three steps: first, a 3D shape is rendered into multiple images using varying camera extrinsics; then image features (e.g. $\text{conv5}$ feature in VGG or AlexNet) are extracted for each view; lastly features are combined across views through a pooling layer, followed by fully connected layers. 


Although the \MVCNN presented by \cite{su15mvcnn} produces compelling results, we are able to improve its performance through a multi-resolution extension with improved data augmentation.
We introduce multi-resolution 3D filtering to capture information at multiple scales. We perform \emph{sphere rendering} (see Sec~\ref{sec:diagnosis}) at different volume resolutions. Note that we use spheres for this discretization as they are view-invariant. 
In particular, this helps regularize out potential noise or irregularities in real-world scanned data (relative to synthetic training data), enabling robust performance on real-world scans. Note that our 3D multi-resolution filtering is different from classical 2D multi-resolution approaches, since the 3D filtering respects the distance in 3D. 

Additionally, we also augment training data with variations in both azimuth and elevation, as opposed to azimuth only. We use AlexNet instead of VGG for efficiency.



\section{Experiments} \label{sec:experiments}

We evaluate our \DCNNs and \MVCNNs along with current state of the art on the ModelNet dataset~\cite{wu20153d} and a new dataset of real-world reconstructions of 3D objects.

For convenience in following discussions, we define \emph{3D resolution} to be the discretization resolution of a 3D shape. That is, a $30\times 30\times 30$ volume has 3D resolution $30$. The sphere rendering from this volume also has 3D resolution $30$, though it may have higher 2D image resolution.

\subsection{Datasets}
\paragraph{ModelNet} 
We use ModelNet~\cite{wu20153d} for our training and testing datasets. ModelNet currently contains $127,915$ 3D CAD models from $662$ categories. ModelNet40, a subset including $12,311$ models from $40$ categories, is well annotated and can be downloaded from the web. The authors also provide a training and testing split on the website, in which there are $9,843$ training and $2,468$ test models\footnote{VoxNet~\cite{maturana2015voxnet} uses the train/test split provided on the website and report average class accuracy on the $2,468$ test split. 3DShapeNets~\cite{wu20153d} and MVCNN~\cite{su15mvcnn} use another train/test split comprising the first 80 shapes of each category in the ``train'' folder (or all shapes if there are fewer than 80) and the first 20 shapes of each category in the ``test'' folder, respectively.}. We  use this train/test split for our experiments. 

By default, we report classification accuracy on all models in the test set (average instance accuracy). For comparisons with previous work we also report average class accuracy.

\begin{figure}
	\centering
	\includegraphics[width=\linewidth]{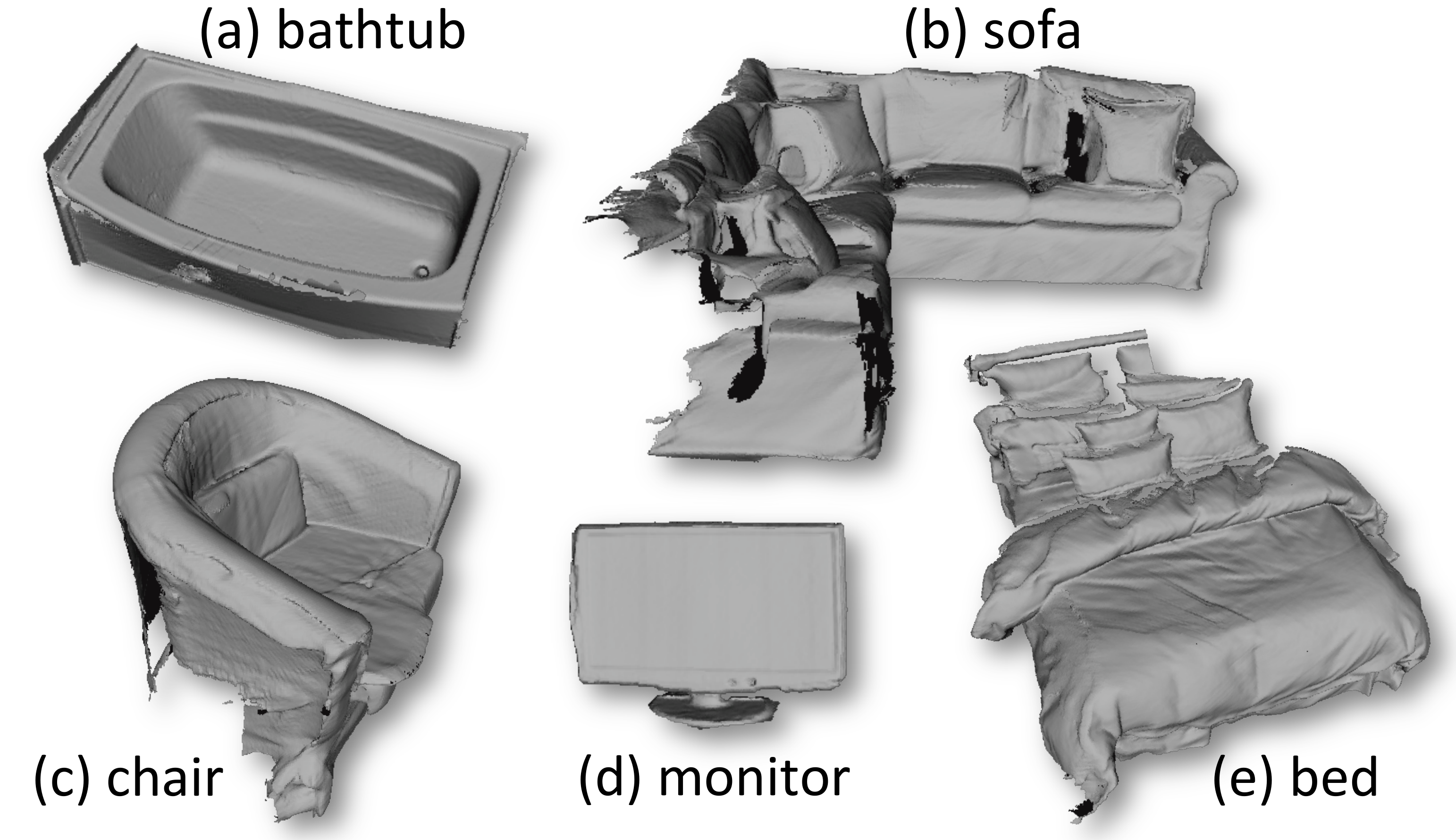}
	\caption{Example models from our real-world dataset. Each model is a dense 3D reconstruction, annotated, and segmented from the background.}
	\label{fig:dataset}
\end{figure}

\paragraph{Real-world Reconstructions} 
We provide a new real-world scanning dataset benchmark, comprising 243 objects of 12 categories; the geometry is captured with an ASUS Xtion Pro and a dense reconstruction is obtained using the publicly-available VoxelHashing framework \cite{niessner2013real}.
For each scan, we have performed a coarse, manual segmentation of the object of interest.
In addition, each scan is aligned with the world-up vector.
While there are existing datasets captured with commodity range sensors -- e.g., \cite{silberman2012indoor,xiao2013sun3d,song2015sun} -- this is the first containing hundreds of annotated models from dense 3D reconstructions.
The goal of this dataset is to provide an example of modern real-time 3D reconstructions; i.e., structured representations more complete than a single RGB-D frame but still with many occlusions. This dataset is used as a test set.

\subsection{Comparison with State-of-the-Art Methods}
\label{sec:exp:comparison_ingroup}
We compare our methods with state of the art for shape classification on the ModelNet40 dataset.
In the following, we discuss the results within \DCNN methods and within \MVCNN methods.

\paragraph{Volumetric CNNs}
\begin{figure}
	\centering
	\includegraphics[width=\linewidth]{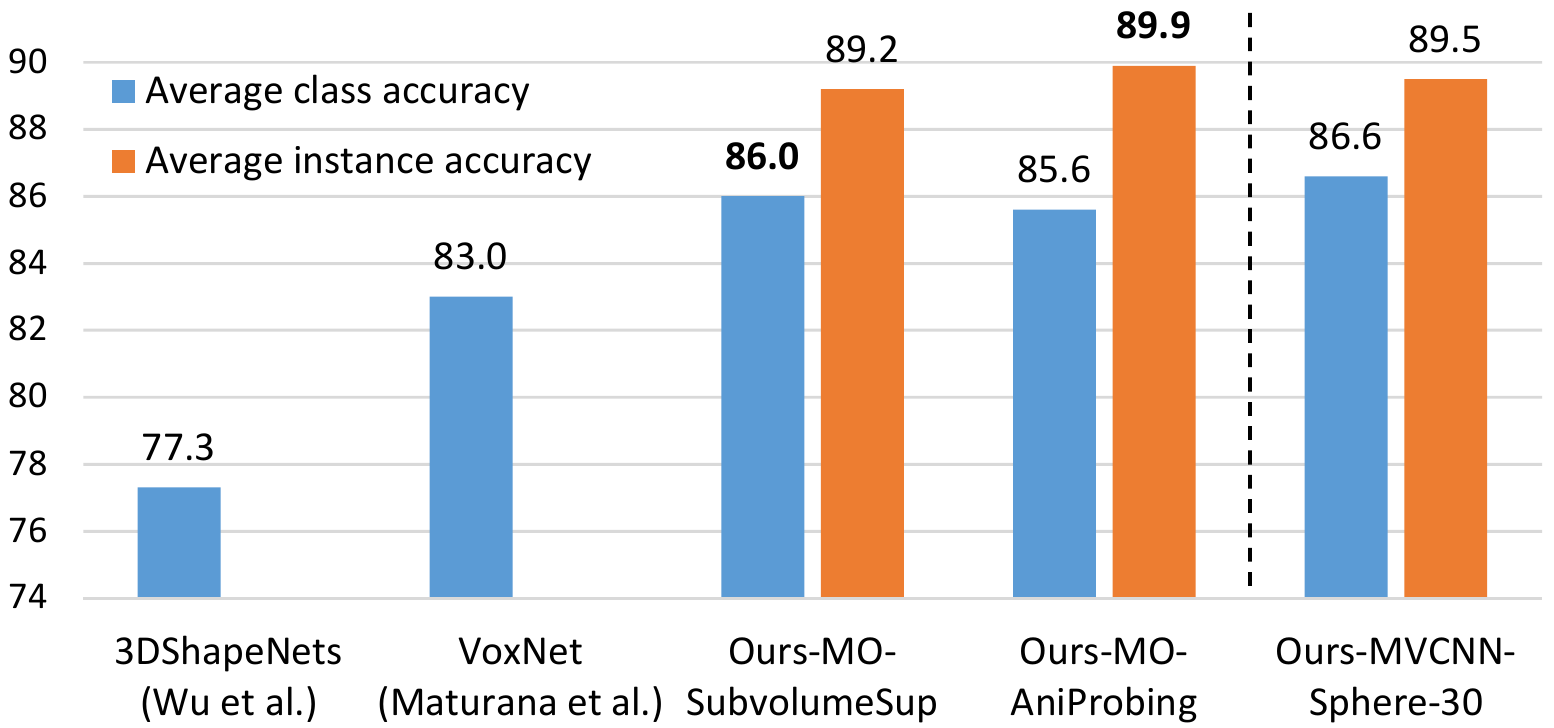}
	\caption{Classification accuracy on ModelNet40 (voxelized at resolution 30). Our \DCNNs have matched the performance of \MVCNN at 3D resolution $30$ (our implementation of \SuMVCNN~\cite{su15mvcnn}, rightmost group).}
	\label{fig:volumetric_results}
\end{figure}
Fig~\ref{fig:volumetric_results} summarizes the performance of \DCNNs. Ours-MO-SubvolumeSup is the subvolume supervision network in Sec~\ref{sec:auxiliary_training} and Ours-MO-AniProbing is the anistropic probing network in Sec~\ref{sec:anisotropic_probing}. Data augmentation is applied as described in Sec~\ref{sec:exp:augmentation} (azimuth and elevation rotations). For clarity, we use MO- to denote that both networks are trained with an additional multi-orientation pooling step ($20$ orientations in practice). For reference of \MVCNN performance at the same 3D resolution, we also include Ours-MVCNN-Sphere-30, the result of our \MVCNN with sphere rendering at 3D resolution $30$. More details of setup can be found in the supplementary. 

As can be seen, both of our proposed volumetric CNNs significantly outperform state-of-the-art volumetric CNNs. Moreover, they both match the performance of our \MVCNN under the same 3D resolution. That is, \emph{the gap between \DCNNs and \MVCNNs is closed} under 3D resolution $30$ on ModelNet40 dataset, an issue that motivates our study (Sec~\ref{sec:diagnosis}). 

\paragraph{Multi-view CNNs}
\begin{figure}
	\centering
	\includegraphics[width=\linewidth]{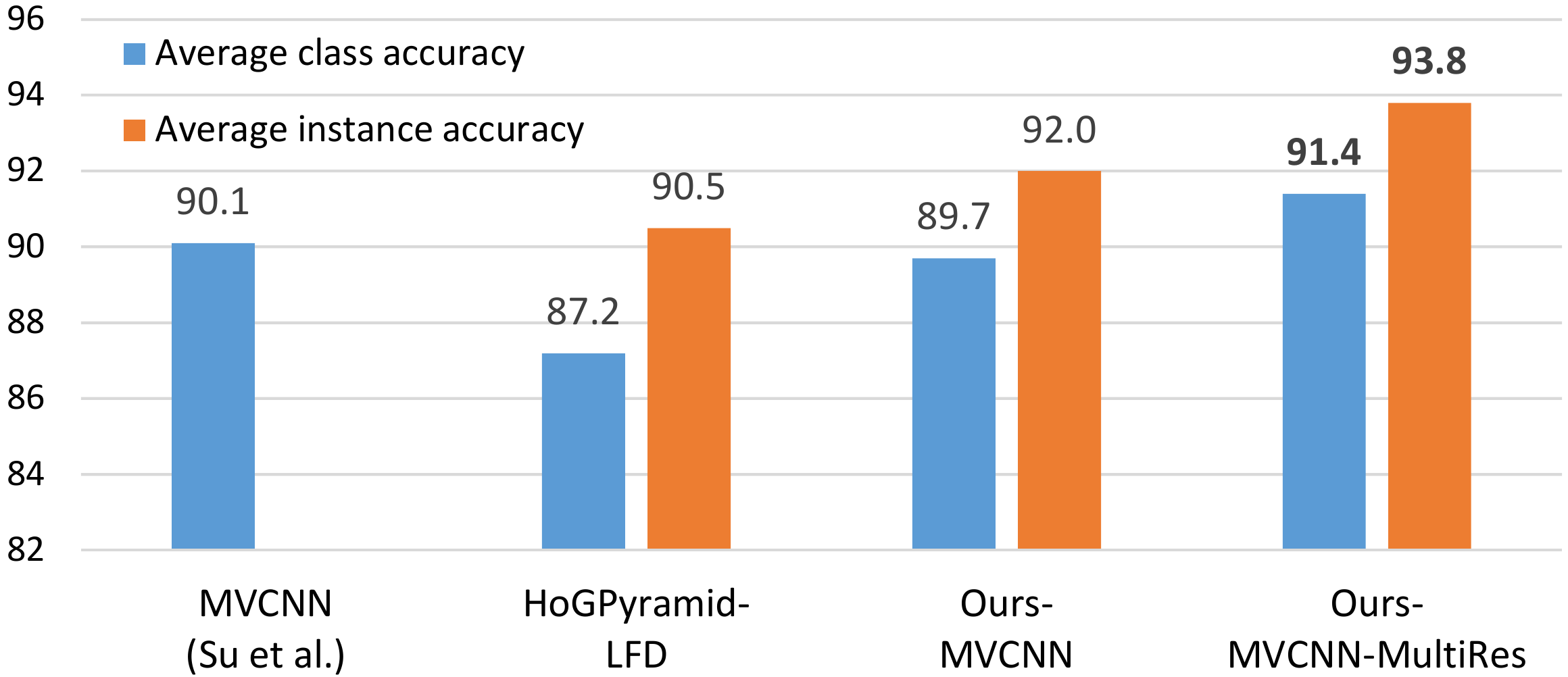}
	\caption{Classification acurracy on ModelNet40 (multi-view representation). The 3D multi-resolution version is the strongest. It is worth noting that the simple baseline HoGPyramid-LFD performs quite well.}
	\label{fig:mvcnn_results}
\end{figure}
Fig~\ref{fig:mvcnn_results} summarizes the performance of \MVCNNs. Ours-MVCNN-MultiRes is the result by training an SVM over the concatenation of fc7 features from Ours-MVCNN-Sphere-30, 60, and Ours-MVCNN. HoGPyramid-LFD is the result by training an SVM over a concatenation of HoG features at three 2D resolutions. Here LFD (lightfield descriptor) simply refers to extracting features from renderings. Ours-MVCNN-MultiRes achieves state-of-the-art.

\subsection{Effect of 3D Resolution over Performance}
\label{sec:exp:comparison_acrossgroup}
Sec~\ref{sec:exp:comparison_ingroup} shows that our \DCNN and \MVCNN performs comparably at 3D resolution $30$. Here we study the effect of 3D resolution for both types of networks. 

\begin{figure}[t!]
	\centering
	\includegraphics[width=\linewidth]{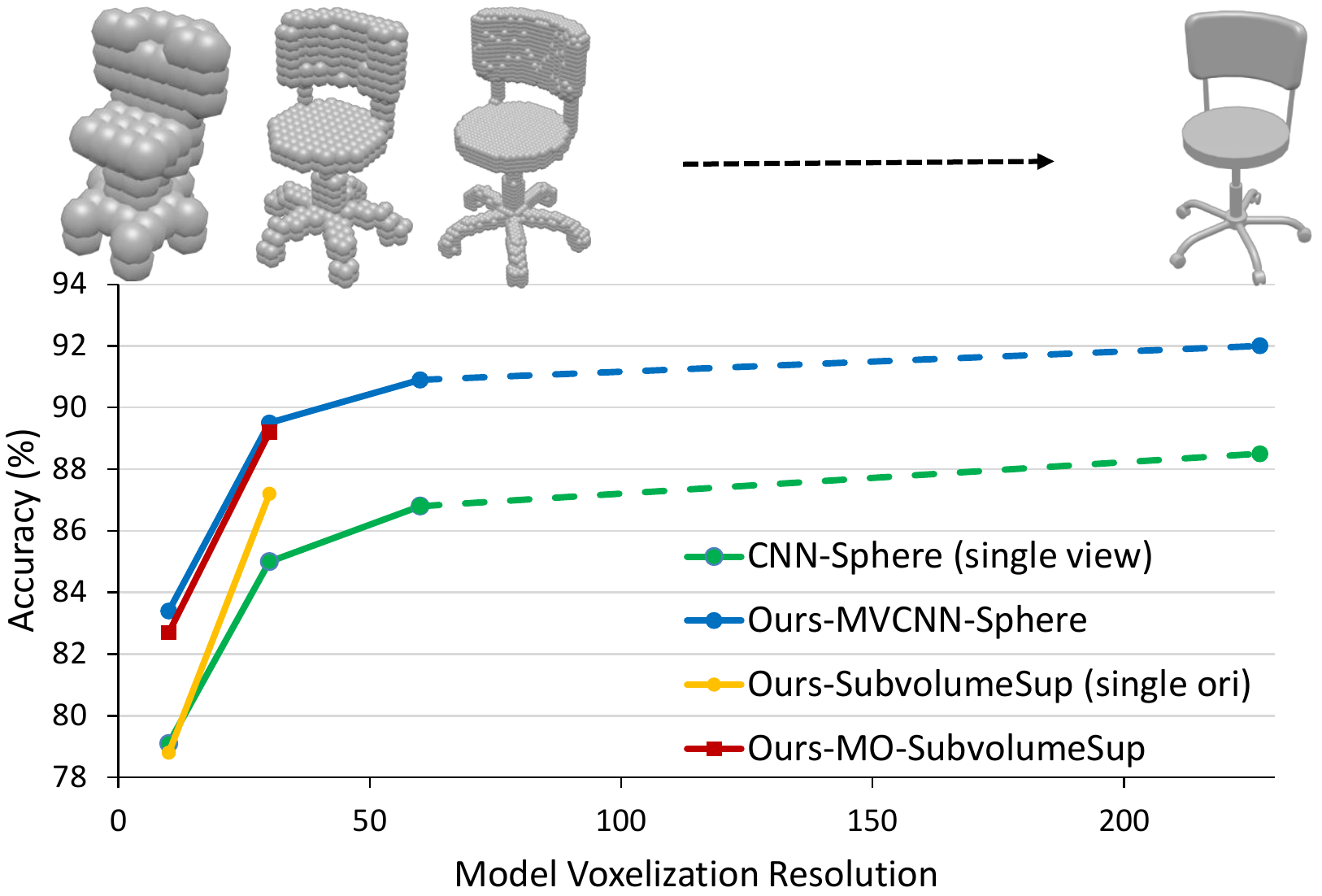}
	\caption{Top: sphere rendering at 3D resolution $10$, $30$, $60$, and standard rendering. Bottom: performance of image-based CNN and volumetric CNN with increasing 3D resolution. The two rightmost points are trained/tested from standard rendering.}
	\label{fig:resolution_mvcnn}
\end{figure}

Fig~\ref{fig:resolution_mvcnn} shows the performance of our \DCNN and \MVCNN at different 3D resolutions (defined at the beginning of Sec~\ref{sec:experiments}). Due to computational cost, we only test our \DCNN at 3D resolutions $10$ and $30$. The observations are: first,  the performance of our \DCNN and \MVCNN is on par at tested 3D resolutions; second, the performance of \MVCNN increases as the 3D resolution grows up. To further improve the performance of \DCNN, this experiment suggests that it is worth exploring how to scale \DCNN to higher 3D resolutions.

\subsection{More Evaluations}
\paragraph{Data Augmentation and Multi-Orientation Pooling}
\label{sec:exp:augmentation}
We use the same \DCNN model, the end-to-end learning verion of 3DShapeNets~\cite{wu20153d}, to train and test on three variations of augmented data (Table~\ref{tb:augmentation}). Similar trend is observed for other volumetric CNN variations.  

\begin{table}[h!]
	\small
	\centering
	\begin{tabular}{lccc}
		\hline
		Data Augmentation & Single-Ori & Multi-Ori & $\Delta$\\
		\hline
		Azimuth rotation (AZ) & $84.7$ & $86.1$ & $1.4$\\
		AZ + translation& $84.8$ & $86.1$ & $1.3$\\
		AZ + elevation rotation & $83.0$ & $\mathbf{87.8}$ & $4.8$\\
		\hline
	\end{tabular}
	\caption{Effects of data augmentations on multi-orientation volumetric CNN. We report numbers of classification accuracy on ModelNet40, with (Multi-Ori) or without (Single-Ori) multi-orientation pooling described in Sec~\ref{sec:arch:aug}.}
	\label{tb:augmentation}
\end{table}
When combined with multi-orientation pooling, applying both azimuth rotation (AZ) and elevation rotation (EL) augmentations is extremely effective. Using only azimuth augmentation (randomly sampled from $0^{\circ}$ to $360^{\circ}$) with orientation pooling, the classification performance is increased by $86.1\%-84.7\%=1.4\%$; combined with elevation augmentation (randomly sampled from $-45^{\circ}$ to $45^{\circ}$), the improvement becomes more significant -- increasing by $87.8\%-83.0\%=4.8\%$. On the other hand, translation jittering (randomly sampled shift from $0$ to $6$ voxels in each direction) provides only marginal influence.


\paragraph{Comparison of Volumetric CNN Architectures}
The architectures in comparison include VoxNet~\cite{maturana2015voxnet},  E2E-\cite{wu20153d} (the end-to-end learning variation of \cite{wu20153d} implemented in Caffe~\cite{jia2014caffe} by ourselves), 3D-NIN (a 3D variation of Network in Network~\cite{lin2013network} designed by ourselves as in Fig~\ref{fig:auxiliary_training} without the ``Prediction by partial object'' branch), SubvolumeSup (Sec~\ref{sec:auxiliary_training}) and AniProbing (Sec~\ref{sec:anisotropic_probing}).  Data augmentation of AZ+EL (Sec~\ref{sec:exp:augmentation}) are applied. 

From Table~\ref{tb:perf},  first, the two \DCNNs we propose, SubvolumeSup and AniProbing networks, both show superior performance, indicating the effectiveness of our design;
second, multi-orientation pooling increases performance for all network variations. This is especially significant for the anisotropic probing network, since each orientation usually only carries partial information of the object. 

\begin{table}
	\small
	\centering
	\begin{tabular}{l|cc}
		\hline
		Network & Single-Ori & Multi-Ori \\
		\hline
		E2E-\cite{wu20153d} & $83.0$ & $87.8$ \\
		VoxNet\cite{maturana2015voxnet} & $83.8$ & $85.9$ \\
		\hline
		3D-NIN & $86.1$ & $88.5$ \\
		Ours-SubvolumeSup & $\mathbf{87.2}$ & $89.2$  \\
		Ours-AniProbing & $85.9$ & $\mathbf{89.9}$  \\
		\hline
	\end{tabular}
	\caption{Comparison of performance of volumetric CNN architectures. Numbers reported are classification accuracy on ModelNet40. Results from E2E-\cite{wu20153d} (end-to-end learning version) and VoxNet~\cite{maturana2015voxnet} are obtained by ourselves. All experiments are using the same set of azimuth and elevation augmented data.}
	\label{tb:perf}
\end{table}

\paragraph{Comparison of Multi-view Methods}

\begin{table}[b!]
	\small
	\centering
	\begin{tabular}{l|ccc}
		\hline
		Method & \#Views & \begin{tabular}[c]{@{}l@{}}Accuracy\\(class)\end{tabular} & \begin{tabular}[c]{@{}l@{}}Accuracy\\(instance)\end{tabular}\\
		\hline
		SPH~(reported by \cite{wu20153d}) & - & $68.2$ & - \\
		LFD~(reported by \cite{wu20153d}) & - & $75.5$ & - \\
		FV (reported by \cite{su15mvcnn}) & 12 & $84.8$ & - \\
		\SuMVCNN~\cite{su15mvcnn} & 80 & $90.1$ & - \\
		\hline
		PyramidHoG-LFD & 20 & $87.2$ & $90.5$\\
		Ours-MVCNN & 20 & $89.7$ & $92.0$\\
	    Ours-MVCNN-MultiRes & 20 & $\mathbf{91.4}$ & $\mathbf{93.8}$ \\
		\hline
	\end{tabular}
	\caption{Comparison of multi-view based methods. Numbers reported are classification accuracy (class average and instance average) on ModelNet40.}
	\label{tb:multiview_performance}
\end{table}

We compare different methods that are based on multi-view representations in Table~\ref{tb:multiview_performance}. Methods in the second group are trained on the full ModelNet40 train set. Methods in the first group, SPH, LFD, FV, and \SuMVCNN, are trained on a subset of ModelNet40 containing 3,183 training samples. They are provided for reference. Also note that the \AbbrMVCNNs in the second group are our implementations in Caffe with AlexNet instead of VGG as in \SuMVCNN~\cite{su15mvcnn}. 

We observe that \AbbrMVCNNs are superior to methods by SVMs on hand-crafted features. 
%
%
%

\paragraph{Evaluation on the Real-World Reconstruction Dataset}

\begin{table}[t!]
	\small
\begin{center}
    \begin{tabular}{l|c|c}
    \hline
    Method      & Classification & Retrieval MAP \\ \hline
    E2E-\cite{wu20153d} & $69.6$ & -\\
    \SuMVCNN~\cite{su15mvcnn} & $72.4$ & $35.8$\\
    \hline
    Ours-MO-SubvolumeSup & $73.3$ & $39.3$\\
    Ours-MO-AniProbing & $70.8$ & $40.2$\\
    Ours-MVCNN-MultiRes & $\mathbf{74.5}$ & $\mathbf{51.4}$\\
    \hline
    \end{tabular}
    \caption{Classification accuracy and retrieval MAP on reconstructed meshes of 12-class real-world scans.}
    \label{tb:real_eval}
\end{center}
\end{table}

We further assess the performance of \DCNNs and \MVCNNs on real-world reconstructions in Table~\ref{tb:real_eval}. All methods are trained on CAD models in ModelNet40 but tested on real data, which may be highly partial, noisy, or oversmoothed~(Fig~\ref{fig:dataset}).
Our networks continue to outperform state-of-the-art results. In particular, our 3D multi-resolution filtering is quite effective on real-world data, possibly because the low 3D resolution component filters out spurious and noisy micro-structures. Example results for object retrieval can be found in supplementary.


\section{Conclusion and Future work}
In this paper, we have addressed the task of object classification on 3D data using \DCNNs and \MVCNNs.
We have analyzed the performance gap between \DCNNs and \MVCNNs from perspectives of network architecture and 3D resolution. 
The analysis motivates us to propose two new architectures of \DCNNs, which outperform state-of-the-art \DCNNs, achieving  comparable performance to \MVCNNs at the same 3D resolution of $30\times 30\times 30$.
Further evalution over the influence of 3D resolution indicates that 3D resolution is likely to be the bottleneck for the performance of \DCNNs. Therefore, it is worth exploring the design of efficient \DCNN architectures that scale up to higher resolutions.

\mypara{Acknowledgement.} The authors gratefully acknowledge the support of Stanford Graduate Fellowship, NSF grants IIS-1528025 and DMS-1546206, ONR MURI grant N00014-13-1-0341, a Google Focused Research award, the Max Planck Center for Visual Computing and Communications and hardware donations by NVIDIA.

{ \small
	\bibliographystyle{ieee}
	\bibliography{main}

\begin{thebibliography}{10}\itemsep=-1pt

\bibitem{bronstein2011shape}
A.~M. Bronstein, M.~M. Bronstein, L.~J. Guibas, and M.~Ovsjanikov.
\newblock Shape google: Geometric words and expressions for invariant shape
  retrieval.
\newblock {\em ACM Transactions on Graphics (TOG)}, 30(1):1, 2011.

\bibitem{chang2015shapenet}
A.~X. Chang, T.~Funkhouser, L.~Guibas, P.~Hanrahan, Q.~Huang, Z.~Li,
  S.~Savarese, M.~Savva, S.~Song, H.~Su, et~al.
\newblock Shapenet: An information-rich 3d model repository.
\newblock {\em arXiv preprint arXiv:1512.03012}, 2015.

\bibitem{chaudhuri2010data}
S.~Chaudhuri and V.~Koltun.
\newblock Data-driven suggestions for creativity support in 3d modeling.
\newblock In {\em ACM Transactions on Graphics (TOG)}, volume~29, page 183.
  ACM, 2010.

\bibitem{chen2003visual}
D.-Y. Chen, X.-P. Tian, Y.-T. Shen, and M.~Ouhyoung.
\newblock On visual similarity based 3d model retrieval.
\newblock In {\em CGF}, volume~22, pages 223--232. Wiley Online Library, 2003.

\bibitem{cimpoi2014describing}
M.~Cimpoi, S.~Maji, I.~Kokkinos, S.~Mohamed, and A.~Vedaldi.
\newblock Describing textures in the wild.
\newblock In {\em CVPR 2014}, pages 3606--3613. IEEE, 2014.

\bibitem{deng2009imagenet}
J.~Deng, W.~Dong, R.~Socher, L.-J. Li, K.~Li, and L.~Fei-Fei.
\newblock Imagenet: A large-scale hierarchical image database.
\newblock In {\em CVPR 2009}, pages 248--255. IEEE, 2009.

\bibitem{donahue2013decaf}
J.~Donahue, Y.~Jia, O.~Vinyals, J.~Hoffman, N.~Zhang, E.~Tzeng, and T.~Darrell.
\newblock Decaf: A deep convolutional activation feature for generic visual
  recognition.
\newblock {\em arXiv preprint arXiv:1310.1531}, 2013.

\bibitem{eitel15iros}
A.~Eitel, J.~T. Springenberg, L.~Spinello, M.~Riedmiller, and W.~Burgard.
\newblock Multimodal deep learning for robust rgb-d object recognition.
\newblock In {\em IEEE/RSJ International Conference on Intelligent Robots and
  Systems (IROS)}, Hamburg, Germany, 2015.

\bibitem{girshick2015fast}
R.~Girshick.
\newblock Fast r-cnn.
\newblock In {\em ICCV 2015}, pages 1440--1448, 2015.

\bibitem{girshick2014rich}
R.~Girshick, J.~Donahue, T.~Darrell, and J.~Malik.
\newblock Rich feature hierarchies for accurate object detection and semantic
  segmentation.
\newblock In {\em CVPR 2014}, pages 580--587. IEEE, 2014.

\bibitem{gupta2014learning}
S.~Gupta, R.~Girshick, P.~Arbel{\'a}ez, and J.~Malik.
\newblock Learning rich features from rgb-d images for object detection and
  segmentation.
\newblock In {\em ECCV 2014}, pages 345--360. Springer, 2014.

\bibitem{han2015matchnet}
X.~Han, T.~Leung, Y.~Jia, R.~Sukthankar, and A.~C. Berg.
\newblock Matchnet: Unifying feature and metric learning for patch-based
  matching.
\newblock In {\em CVPR 2015}, pages 3279--3286, 2015.

\bibitem{horn1984extended}
B.~K. Horn.
\newblock Extended gaussian images.
\newblock {\em Proceedings of the IEEE}, 72(12):1671--1686, 1984.

\bibitem{ioffe2015batch}
S.~Ioffe and C.~Szegedy.
\newblock Batch normalization: Accelerating deep network training by reducing
  internal covariate shift.
\newblock {\em arXiv preprint arXiv:1502.03167}, 2015.

\bibitem{jaderberg2015spatial}
M.~Jaderberg, K.~Simonyan, A.~Zisserman, et~al.
\newblock Spatial transformer networks.
\newblock In {\em Advances in Neural Information Processing Systems}, pages
  2008--2016, 2015.

\bibitem{jia2014caffe}
Y.~Jia, E.~Shelhamer, J.~Donahue, S.~Karayev, J.~Long, R.~Girshick,
  S.~Guadarrama, and T.~Darrell.
\newblock Caffe: Convolutional architecture for fast feature embedding.
\newblock {\em arXiv preprint arXiv:1408.5093}, 2014.

\bibitem{kazhdan2003rotation}
M.~Kazhdan, T.~Funkhouser, and S.~Rusinkiewicz.
\newblock Rotation invariant spherical harmonic representation of 3 d shape
  descriptors.
\newblock In {\em SGP 2003}, volume~6, pages 156--164, 2003.

\bibitem{knopp2010hough}
J.~Knopp, M.~Prasad, G.~Willems, R.~Timofte, and L.~Van~Gool.
\newblock Hough transform and 3d surf for robust three dimensional
  classification.
\newblock In {\em ECCV 2010}, pages 589--602. Springer, 2010.

\bibitem{kokkinos2012intrinsic}
I.~Kokkinos, M.~M. Bronstein, R.~Litman, and A.~M. Bronstein.
\newblock Intrinsic shape context descriptors for deformable shapes.
\newblock In {\em CVPR 2012}, pages 159--166. IEEE, 2012.

\bibitem{krizhevsky2012imagenet}
A.~Krizhevsky, I.~Sutskever, and G.~E. Hinton.
\newblock Imagenet classification with deep convolutional neural networks.
\newblock In {\em Advances in neural information processing systems}, pages
  1097--1105, 2012.

\bibitem{lecun1998gradient}
Y.~LeCun, L.~Bottou, Y.~Bengio, and P.~Haffner.
\newblock Gradient-based learning applied to document recognition.
\newblock {\em Proceedings of the IEEE}, 86(11):2278--2324, 1998.

\bibitem{lecun2004learning}
Y.~LeCun, F.~J. Huang, and L.~Bottou.
\newblock Learning methods for generic object recognition with invariance to
  pose and lighting.
\newblock In {\em CVPR 2014}, volume~2, pages II--97. IEEE, 2004.

\bibitem{lin2013network}
M.~Lin, Q.~Chen, and S.~Yan.
\newblock Network in network.
\newblock {\em arXiv preprint arXiv:1312.4400}, 2013.

\bibitem{maturana2015voxnet}
D.~Maturana and S.~Scherer.
\newblock Voxnet: A 3d convolutional neural network for real-time object
  recognition.
\newblock In {\em IEEE/RSJ International Conference on Intelligent Robots and
  Systems}, September 2015.

\bibitem{niessner2013real}
M.~Nie{\ss}ner, M.~Zollh{\"o}fer, S.~Izadi, and M.~Stamminger.
\newblock Real-time 3d reconstruction at scale using voxel hashing.
\newblock {\em ACM Transactions on Graphics (TOG)}, 32(6):169, 2013.

\bibitem{osada2002shape}
R.~Osada, T.~Funkhouser, B.~Chazelle, and D.~Dobkin.
\newblock Shape distributions.
\newblock {\em ACM Transactions on Graphics (TOG)}, 21(4):807--832, 2002.

\bibitem{razavian2014cnn}
A.~S. Razavian, H.~Azizpour, J.~Sullivan, and S.~Carlsson.
\newblock Cnn features off-the-shelf: an astounding baseline for recognition.
\newblock In {\em Computer Vision and Pattern Recognition Workshops (CVPRW),
  2014 IEEE Conference on}, pages 512--519. IEEE, 2014.

\bibitem{shi2015deeppano}
B.~Shi, S.~Bai, Z.~Zhou, and X.~Bai.
\newblock Deeppano: Deep panoramic representation for 3-d shape recognition.
\newblock {\em Signal Processing Letters, IEEE}, 22(12):2339--2343, 2015.

\bibitem{silberman2012indoor}
N.~Silberman, D.~Hoiem, P.~Kohli, and R.~Fergus.
\newblock Indoor segmentation and support inference from rgbd images.
\newblock In {\em ECCV 2012}, pages 746--760. Springer, 2012.

\bibitem{socher2012convolutional}
R.~Socher, B.~Huval, B.~Bath, C.~D. Manning, and A.~Y. Ng.
\newblock Convolutional-recursive deep learning for 3d object classification.
\newblock In {\em NIPS 2012}, pages 665--673, 2012.

\bibitem{song2015sun}
S.~Song, S.~P. Lichtenberg, and J.~Xiao.
\newblock Sun rgb-d: A rgb-d scene understanding benchmark suite.
\newblock In {\em CVPR 2015}, pages 567--576, 2015.

\bibitem{su15mvcnn}
H.~Su, S.~Maji, E.~Kalogerakis, and E.~G. Learned{-}Miller.
\newblock Multi-view convolutional neural networks for 3d shape recognition.
\newblock In {\em ICCV 2015}, 2015.

\bibitem{wu20153d}
Z.~Wu, S.~Song, A.~Khosla, F.~Yu, L.~Zhang, X.~Tang, and J.~Xiao.
\newblock 3d shapenets: A deep representation for volumetric shapes.
\newblock In {\em CVPR 2015}, pages 1912--1920, 2015.

\bibitem{xiao2013sun3d}
J.~Xiao, A.~Owens, and A.~Torralba.
\newblock Sun3d: A database of big spaces reconstructed using sfm and object
  labels.
\newblock In {\em ICCV 2013}, pages 1625--1632. IEEE, 2013.

\end{thebibliography}
}

\newpage

\appendix
\section{Appendix}
In this section, we present positive effects of two adds-on modules -- volumetric batch normalization (Sec~\ref{sec:batchnorm}) and spatial transformer networks (Sec~\ref{sec:stn}). We also provide more details on experiments in the main paper (Sec~\ref{sec:training_details}) and real-world dataset construction (Sec~\ref{sec:realworld}). Retrieval results can also be found in Sec~\ref{sec:retrieval_results}.

\subsection{Batch Normalization}
\label{sec:batchnorm}
We observe that using batch normalization~\cite{ioffe2015batch} can accelerate the training process and also improve final performance. Taking our subvolume supervision model (base network is 3D-NIN) for example, the classification accuracy from single orientation is $87.2\%$ and $88.8\%$ before and after using batch normalization, respectively. Complete results are in Table~\ref{tab:bn}.

Specifically, compared with the model described in the main paper, we add batch normalization layers after each convolution and fully connected layers. We also add dropout layers after each convolutional layers.

\begin{table}[h]
	\small
	\centering
	\begin{tabular}{l|cccc}
		\hline
		Model & Single-Ori & Multi-Ori \\
		\hline
		Ours-SubvolSup & $87.2$ & $89.2$  \\
		Ours-AniProbing & $85.9$ & $89.9$  \\
		\hline
		Ours-SubvolSup + BN & $88.8$ & $90.1$ \\
		Ours-AniProbing + BN & $87.5$ & $90.0$ \\
		\hline
	\end{tabular}
	\caption{Positive effect of adding batch normalization at convolutional layers. Numbers reported are classification (instace average) on ModelNet40 test set.}
	\label{tab:bn}
\end{table}

\subsection{Spatial Transformer Networks}
\label{sec:stn}
One disadvantage of multi-view/orientation method is that one needs to prepare multiple views/orientations of the 3D data, thus computationally more expensive. It would be ideal if we can achieve similar performance with just a single input. In this section we show how a Spatial Transformer Network (STN)~\cite{jaderberg2015spatial} can help boost our model's performance on single-orientation input. 

\begin{table}[h]
	\small
	\centering
	\begin{tabular}{l|c}
		\hline
		Model & Single-Ori \\
		\hline
		Ours-SubvolSup + BN & $88.8$ \\
		Ours-SubvolSup + BN + STN & $89.1$ \\
		\hline
	\end{tabular}
	\caption{Spatial transformer network helps improve single orientation classification accuracy.}
	\label{tab:stn}
\end{table}

\begin{figure}[h]
	\centering
	\includegraphics[width=0.75\linewidth]{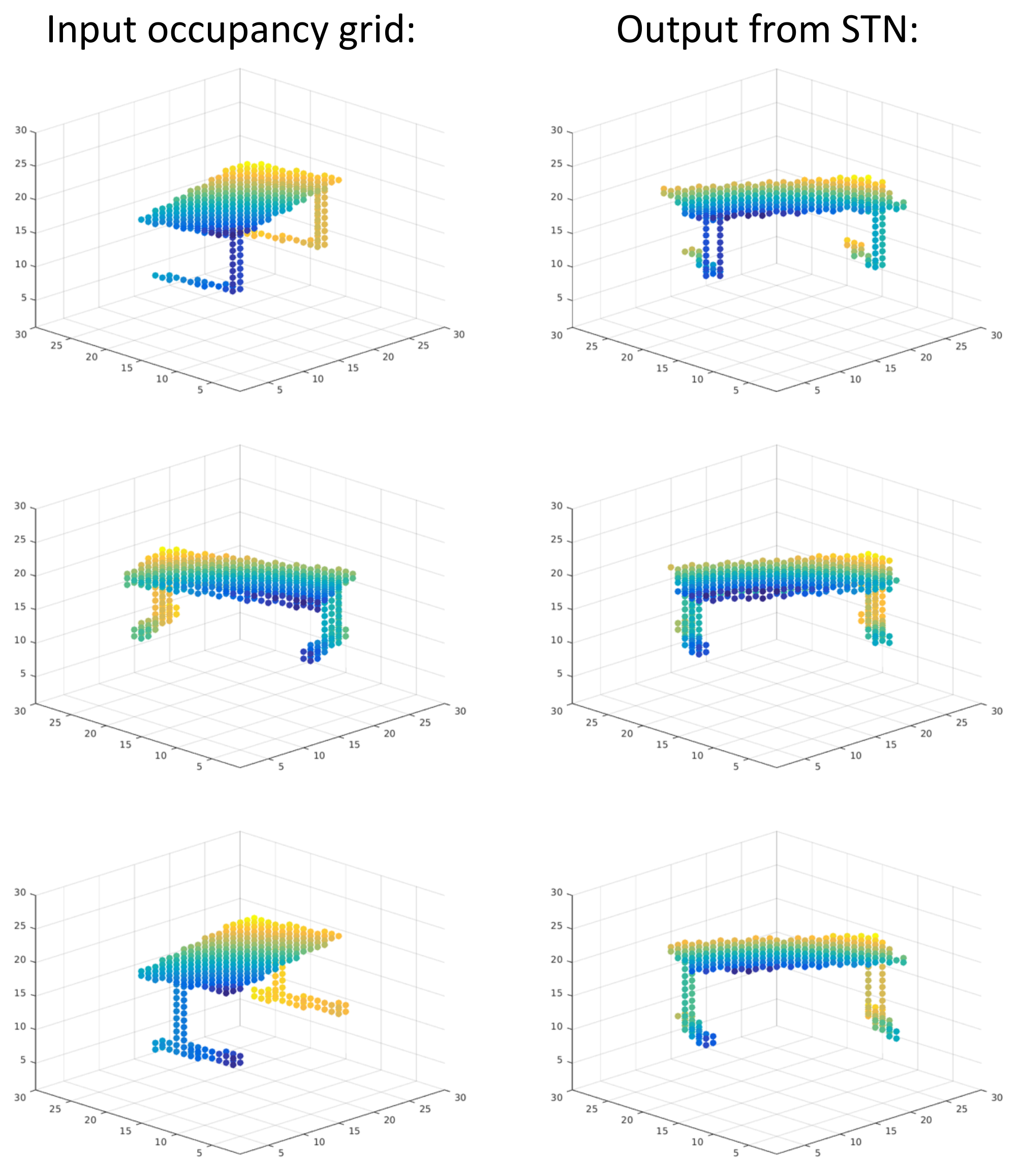}
	\caption{Each row is a input and output pair of the spatial transformer netowrk (`table' category). Each point represents an occupied voxel and color is determined by depth. We see STN tends to align all the tables to a canonical viewpoint.}
	\label{fig:stn}
\end{figure}

The spatial transformer network has three components: (1) a regressor network which takes occupancy grid as input and predicts transformation parameters. (2) a grid generator that outputs a sampling grid based on the transformation and (3) a sampler that transforms the input volume to a new volume based on the sampling grid. We include a spatial transfomer network directly after the data layer and before the original volumetric CNN (see Table~\ref{tab:stn} for results). In Fig~\ref{fig:stn}, we visualize the effect of spatial transformer network on some exemplar input occupancy grids.

\subsection{Details on Model Training}
\label{sec:training_details}

\paragraph{Training for Our Volumetric CNNs} To produce occupancy grids from meshes, the faces of a mesh are subdivided until the length of the longest edge is within a single voxel; then all voxels that intersect with a face are marked as occupied. For 3D resolution 10,30 and 60 we generate voxelizations with central regions $10$, $24$, $54$ and padding $0$, $3$, $3$ respectively.

This voxelization is followed by a hole filling step that fills the holes inside the models as occupied voxels.

To augment our training data with azimuth and elevation rotations, we generate 60 voxelizations for each model, with azimuth uniformly sampled from $[0,360]$ and elevation uniformly sampled from $[-45,45]$ (both in degrees). 

We use a Nesterov solver with learning rate $0.005$ and weight decay $0.0005$ for training. It takes around 6 hours to train on a K40 using Caffe~\cite{jia2014caffe} for the subvolume supervision CNN and 20 hours for the anisotropic probing CNN. For multi-orientation versions of them, SubvolumeSup splits at the last conv layer and AniProbing splits at the second last conv layer. Volumetric CNNs trained on single orientation inputs are then used to initialize their multi-orientation version for fine tuning.

During testing time, 20 orientations of a CAD model occupancy grid (equally distributed azimuth and uniformly sampled elevation from  $[-45,45]$) are input to MO-VCNN to make a class prediction.

\paragraph{Training for Our MVCNN and Multi-resolution MVCNN} We use Blender to render 20 views of each (either ordinary or spherical) CAD model from azimuth angles in $0,36,72,...,324$ degrees and elevation angles in $-30$ and $30$ degrees. For sphere rendering, we convert voxelized CAD models into meshes by replacing each voxel with an approximate sphere with 50 faces and diameter length of the voxel size. Four fixed point light sources are used for the ray-tracing rendering.

We first finetune AlexNet with rendered images for ordinary rendering and multi-resolutional sphere renderings separately. Then we use trained AlexNet to initialize the MVCNN and fine tune on multi-view inputs.


\paragraph{Other Volumetric Data Representations}
Note that while we present our \DCNN methods using occupancy grid representations of 3D objects, our approaches easily generalize to other volumetric data representations. 
In particular, we have also used Signed Distance Functions and (unsigned) Distance Functions as input (also $30\times 30\times 30$ grids). Signed distance fields were generated through virtual scanning of synthetic training data, using volumetric fusion (for our real-world reconstructed models, this is the natural representation); distance fields were generated directly from the surfaces of the models.
Performance was not affected significantly by the different representations, differing by around $0.5\%$ to $1.0\%$ for classification accuracy on ModelNet test data.

\begin{figure*}[t!]
\centering
\includegraphics[width=\linewidth]{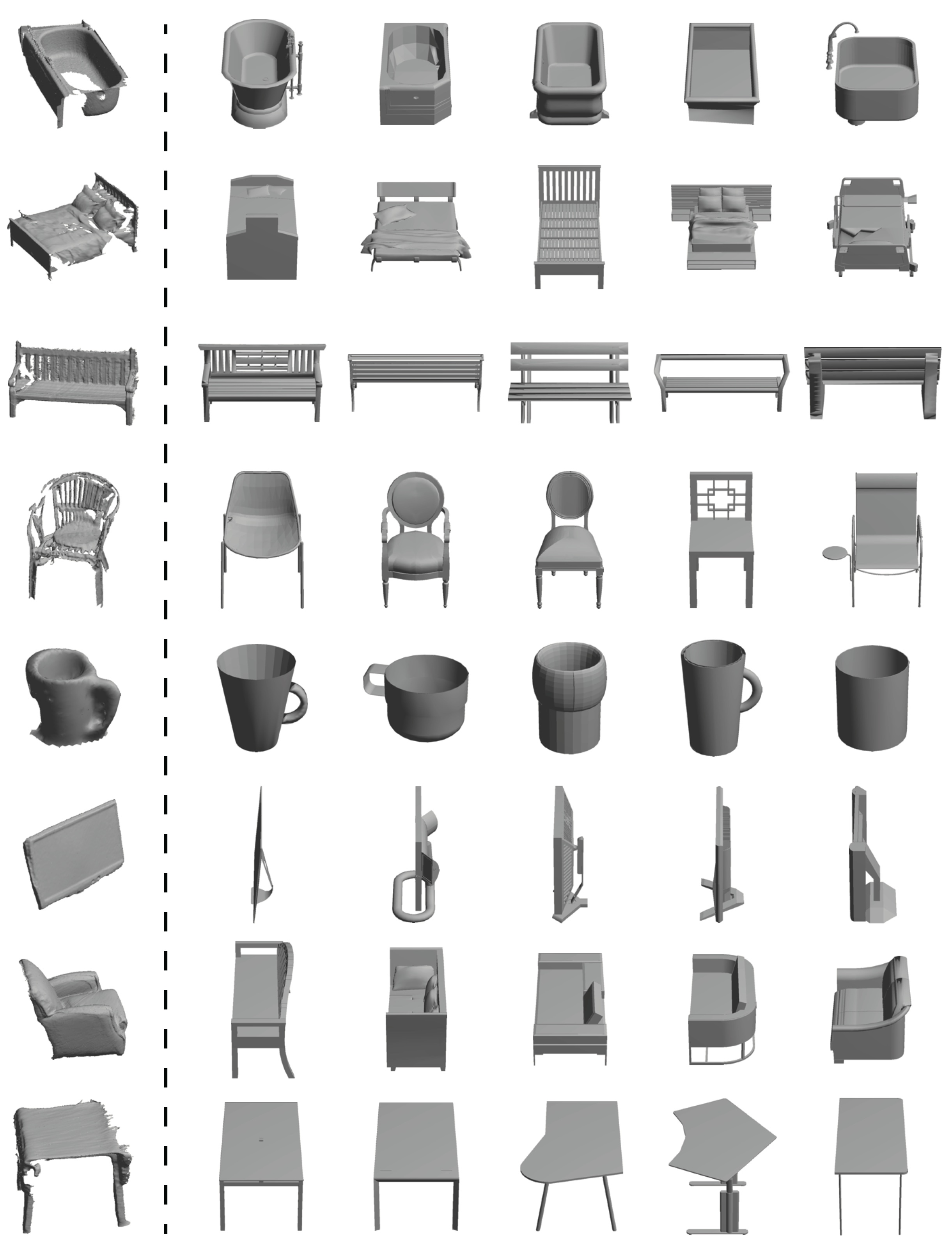}
\caption{More retrieval results. Left column: queries, real reconstructed meshes. Right five columns: retrieved models from ModelNet40 Test800.}
\label{fig:retrieval_1}
\end{figure*}

\begin{figure*}[t!]
\centering
\includegraphics[width=\linewidth]{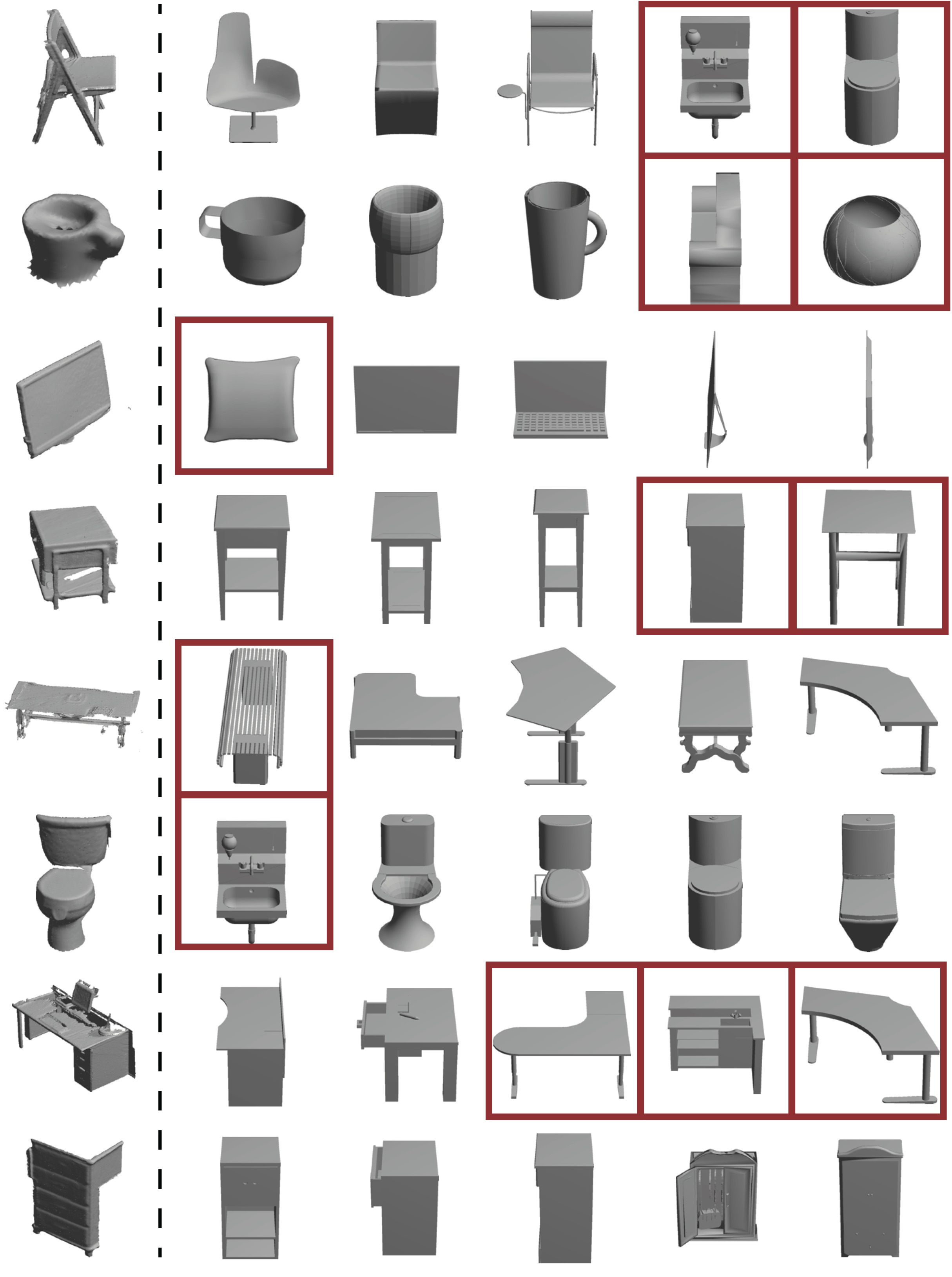}
\caption{More retrieval results (samples with mistakes). Left column: queries, real reconstructed meshes. Right five columns: retrieved models from ModelNet40 Test800. Red bounding boxes denote results from wrong categories.}
\label{fig:retrieval_2}
\end{figure*}

\subsection{Real-world Reconstruction Test Data}
\label{sec:realworld}

\begin{figure*}[t!]
\centering
\includegraphics[width=\linewidth]{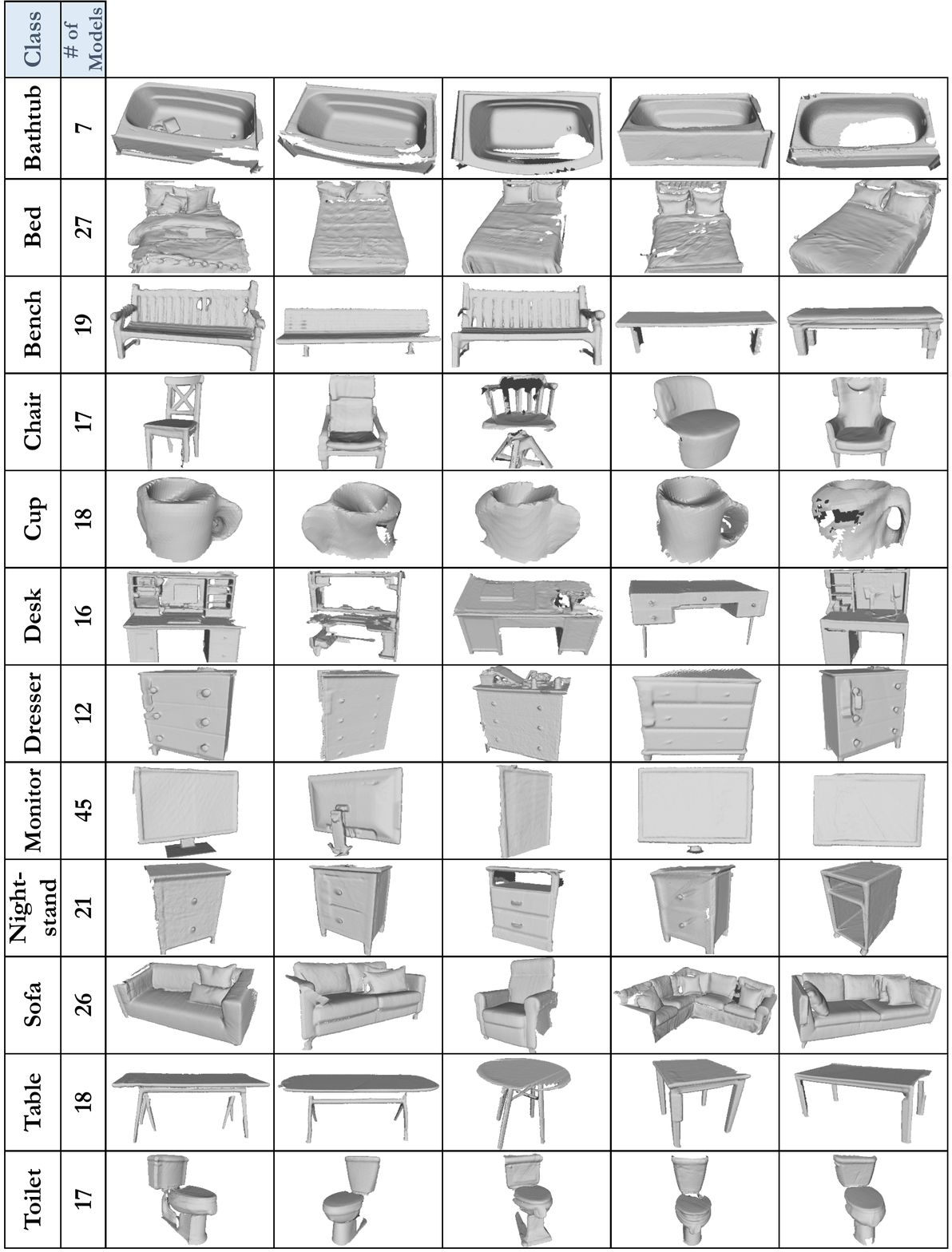}
\caption{Our real-world reconstruction test dataset, comprising 12 categories and 243 models. Each row lists a category along with the number of objects and several example reconstructed models in that category.}
\label{fig:test_dataset}
\end{figure*}



In order to evaluate our method on real scanning data, we obtain a dataset of 3D models, which we reconstruct using data from a commodity RGB-D sensor (ASUS Xtion Pro). 
To this end, we pick a variety of real-world objects for which we record a short RGB-D frame sequence (several hundred frames) for each instance.
For each object, we use the publicly-available Voxel Hashing framework in order to obtain a dense 3D reconstruction. 
In a semi-automatic post-processing step, we segment out the object of interest's geometry by removing the scene background. 
In addition, we align the obtained model with the world up direction. 
Overall, we obtained scans of 243 objects, comprising of a total of over XYZ thousand RGB-D input frames.

\subsection{More Retrieval Results}
\label{sec:retrieval_results}
For model retrieval, we extract CNN features (either from 3D CNNs or MVCNNs) from query models and find nearest neighbor results based on L2 distance. Similar to MVCNN (Su et al.)~\cite{su15mvcnn}, we use a low-rank Mahalanobis metric to optimize retrieval performance.
Figure~\ref{fig:retrieval_1} and Figure~\ref{fig:retrieval_2} show more examples of retrieval from real model queries.

\end{document}